%% file: DNAFragMimic.tex
\documentclass{WileyMSP-template}
\usepackage{todonotes}
\usepackage{subcaption}
\usepackage{graphicx}
\begin{document}

\pagestyle{fancy}
%\rhead{\includegraphics[width=2.5cm]{vch-logo.png}}
\rhead{\includegraphics[width=1.0cm]{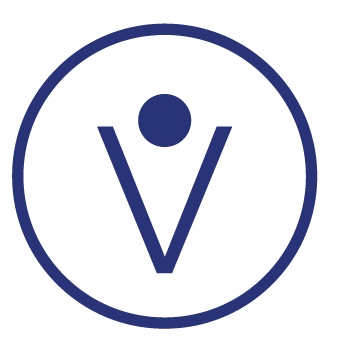}}

%\title{On the non-destructive replacement of chemical assays using machine learning and image processing for sperm DNA fragmentation}
%\title{Mimicking Chemical Assays for Sperm DNA Fragmentation Detection with Machine Learning}
%\title{Replicating Chemical Tests Using Machine Learning for Digital Sperm Analysis}
%\title{Stain-free prediction of DNA fragmentation in human sperm using by machine learning}
%\title{Predicting DNA fragmentation: A non-destructive, stain free approach using machine learning}
\title{Predicting DNA fragmentation: A non-destructive analogue to chemical assays using machine learning}
\maketitle

%\listoftodos
% Author: Please give full first and last names for authors and include * after the name of all corresponding authors

\author{Byron A Jacobs*}
\author{Ifthakaar Shaik}
\author{Frando Lin}

% Dedication

\dedication{}

% Affiliations: Please provide adacemic titles (Prof. or Dr.) for all authors where applicable, and include an institutional email address for all corresponding authors
\begin{affiliations}
Prof B. A. Jacobs\\
VitruvianMD Pte Ltd, 3 Fraser Street, 189352, Singapore, Singapore\\
University of Johannesburg, Department of Mathematics and Applied Mathematics, \\
Johannesburg, Gauteng, South Africa\\
Email Address: byron.jacobs@vitruvianmd.com, byronj@uj.ac.za\\

I. Shaik, F. Lin\\
VitruvianMD Pte Ltd, 3 Fraser Street, 189352, \\
Singapore, Singapore\\
\end{affiliations}

% Keywords: Please provide a minimum of three and a maximum of seven keywords, separated by commas

\keywords{DNA Integrity, Sperm Morphology, Machine Learning, Ensemble}

% Abstract should be written in the present tense and impersonal style (i.e., avoid we), and be at most 200 words long
\begin{abstract}
Globally, infertility rates are increasing, with 2.5\% of all births being assisted by in vitro fertilisation (IVF) in 2022. Male infertility is the cause for approximately half of these cases. The quality of sperm DNA has substantial impact on the success of IVF. The assessment of sperm DNA is traditionally done through chemical assays which render sperm cells ineligible for IVF. Many compounding factors lead to the population crisis, with fertility rates dropping globally in recent history. As such assisted reproductive technologies (ART) have been the focus of recent research efforts. Simultaneously, artificial intelligence has grown ubiquitous and is permeating more aspects of modern life. With the advent of state-of-the-art machine learning and its exceptional performance in many sectors, this work builds on these successes and proposes a novel framework for the prediction of sperm cell DNA fragmentation from images of unstained sperm. Rendering a predictive model which preserves sperm integrity and allows for optimal selection of sperm for IVF.
\end{abstract}

% Text: Please use section headings and subheadings as specified below. For communications, all section headings apart from Experimental Section should be removed
% Please make the first reference to a display item bold: \textbf{Figure 1}
% Do not abbreviate Figure, Equation, etc.; display items are always singular, i.e., Figure 1 and 2.
% Equations are always singular, i.e., Equation 1 and 2, and should be inserted using the {equation} environment, not as graphics
% Please do not use footnotes in the text, additional information can be added to the Reference list.

\input{tex/introduction.tex}
\input{tex/methods.tex}
\input{tex/results.tex}
\input{tex/conclusion.tex}

% Acknowledgements
\medskip
\textbf{Acknowledgements} \par %delete if not applicable))
The authors acknowledge A. Morris, M. Setumo, N. Bassa, M.I. Cassim, Y.M. Dasoo, G. Boshoff and C. Huyser for their expertise and contributions in assay preparation and image capturing, as well as BioART Fertility Clinic and Reproductive Biology Lab of University of Pretoria for their laboratory support. Additionally we acknowledge J. Kruger and S. Frohlich for their assistance in software development and technical support.

% References
\medskip

% Use the following code if you wish to generate your bibliography with BibTeX;
% replace the string "MSP-template" below with the name(s) of
% the BibTeX data base(s) you want to use.
% The resulting bibliography-output (the content of the .bbl file)
% must be pasted back into this file before submission.
% Please also include your BibTeX data base file(s) in your submission
% so that we can re-run BibTeX if necessary.
%

\bibliographystyle{MSP}
\bibliography{DNAFragMimic.bib}

%
%\begin{figure}
%  \includegraphics[width=\linewidth]{placeholder-image.png}
%  \caption{Figure 1 caption goes here. Reproduced with permission.\textsuperscript{[Ref.]} Copyright Year, Publisher. }
%  \label{fig:boat1}
%\end{figure}
%
%\begin{figure}
%  \includegraphics[width=\linewidth]{placeholder-image.png}
%  \caption{Figure 2 caption goes here. Reproduced with permission.\textsuperscript{[Ref.]} Copyright Year, Publisher.}
%  \label{fig:boat1}
%\end{figure}
%
%\begin{figure}
%  \includegraphics[width=\linewidth]{placeholder-image.png}
%  \caption{Figure 3 caption goes here. Reproduced with permission.\textsuperscript{[Ref.]} Copyright Year, Publisher.}
%  \label{fig:boat1}
%\end{figure}
%
%\begin{table}
% \caption{Table 1 caption}
%  \begin{tabular}[htbp]{@{}lll@{}}
%    \hline
%    Description 1 & Description 2 & Description 3 \\
%    \hline
%    Row 1, Col 1  & Row 1, Col 2  & Row 1, Col 3  \\
%    Row 2, Col 1  & Row 2, Col 2  & Row 2, Col 3  \\
%    \hline
%  \end{tabular}
%\end{table}
%

\end{document}

%% file: tex/introduction.tex
\section{Introduction}
%% Background

%Talk about assays, how they agree and disagree. But are destructive.
%Talk about how morphology and motiltiy have been found to correlate with DNA frag
According to a recent study published by the WHO 1-in-6 couples are affected by infertility, where infertility is defined as a disease of the male or female reproductive system resulting in a failure to achieve a pregnancy after 12 months or more of regular unprotected sexual intercourse \cite{world2023infertility}. This study also indicates that infertility rates between high income and middle-to-low income countries are similar. Furthermore, Borumandnia et al. \cite{borumandnia2021assessing} highlight the rising infertility rates across all countries. With a global infertility crisis looming, the need for technological innovation is more pertinent than ever.\\\vspace*{0.5cm}

In approximately one third of infertile couples the problem is male infertility \cite{chandra2013infertility}. It is the focus of this work to develop an automated sperm analysis tool which enables the prediction of sperm DNA integrity from bright field and phase contrast images of unstained sperm. This work is based on data collected in partnership with the Reproductive Biology Lab (RBL) at the University of Pretoria. The data collected comprises samples collected from 65 consenting males who met the inclusion criteria, each of these samples were analysed by a variety of chemical assays, used for measuring the DNA integrity of human sperm.\\\vspace*{0.5cm}

The chemical assays used to generate image data are; Aniline Blue (AB), Toluidine Blue (TB) \cite{pourmasumi2019evaluation}, Acridine Orange \cite{ajina2017assessment}, Chromomycin A3 (CMA3) \cite{zandemami2012correlation}, terminal deoxynucleotidyl transferase-mediated deoxyuridine triphosphate-nick end labelling (TUNEL) \cite{sharma2021tunel} and Sperm Chromatin Dispersion (SCD), specifically GoldCyto \cite{pratap2017assessment} . Additionally, all patient samples were imaged for morphological analysis using Diff-Quick \cite{lincoln2023comparison} and finally, images of unstained samples were captured.\\\vspace*{0.5cm}

The array of assays used in this study measure DNA integrity of sperm using a variety of mechanisms. As such, we endeavour to produce a predictive model for each assay and correlate the results from our machine learning enabled framework with the results generated by the respective chemical assay. It is beyond the scope of this work to cross correlate the chemical tests with one another. Furthermore, we note that all of the chemical assays used to measure DNA fragmentation render the sperm cells unviable for fertilisation. This forms the base premise for our proposed framework, to allow for the prediction of sperm DNA integrity in a non-destructive manner, using image analysis and machine learning, enabling sperm specific selection guidance for assisted fertilisation. \\\vspace*{0.5cm}

According to \cite{pourmasumi2019evaluation} the mechanism of AB is able to differentiate between lysine-rich histones and arginine/cysteine-rich protamines which enables a specific positive reaction for lysine and reveals differences in the basic nuclear protein composition of human sperm. While TB is used for metachromatic staining of the chromatin. The phosphate residues of sperm DNA in nuclei with loosely packed chromatin or damaged DNA are more likely to bind with the basic TB dye, causing a metachromatic shift from light blue to purple–violet due to the dimerization of dye molecules.\\\vspace*{0.5cm}

The AO test is a well-established cytochemical technique used to assess sperm DNA integrity. It differentiates between normal double-stranded DNA and abnormal denatured or single-stranded DNA by leveraging the metachromatic properties of the stain \cite{ajina2017assessment}. Using this differentiation one is able to measure the relative abundances of double and single stranded DNA present in a sperm cell. McCallum et al. \cite{mccallum2019deep} leverage this to train a deep learning framework for selecting sperm with high DNA integrity based on the AO test. \\\vspace*{0.5cm}

CMA3 can be used as an indirect method for assessing sperm protamine \cite{zandemami2012correlation}, showing correlation with direct measures of protamine using, the more time consuming and complex, electrophoresis. \\\vspace*{0.5cm}

TUNEL has become a popular and ubiquitous method for assessing sperm DNA integrity, as it directly measures breaks in both single and double stranded DNA \cite{sharma2021tunel}. \\\vspace*{0.5cm}

Finally, the SCD based tests detects sperm DNA fragmentation by incubating the sperm cells in an acid solution which denatures the DNA. This is followed by the use of a lysing solution that removes the majority of nuclear proteins. The result is the spreading out of DNA loops which produce halos around the cells \cite{fernandez2003sperm}. Due to the destructive nature of this test, the original sperm morphology is entirely lost. \\\vspace*{0.5cm}

To the best of the authors' knowledge, the above data collection procedure presents the most comprehensive cross-sectional study of DNA fragmentation techniques for sperm analysis to be undertaken by a single lab. This is highly beneficial in that the patient samples are consistent across all tests, the method of preparation is consistent, the interpretation of the staining is consistent, unstained and morphological data is also captured per patient. \\\vspace*{0.5cm}

The correlation between sperm morphological and motile health and DNA fragmentation is well established in the literature. Belloc et al. \cite{belloc2014isolated} investigated the correlation between semen parameters and DNA damage, concluding that the highest level of DNA fragmentation was associated with sperm exhibiting motility defects. This finding is corroborated in a study by Le et al. \cite{le2019does} in which the researchers found that DFI was significantly correlated with (positively) abnormal head morphology and (weakly negatively) with progressive motility. Further to the above findings, de la Calle et al. \cite{de2008sperm} showed statistically significant correlation between sperm DNA fragmentation rate and sperm morphology, motility and concentration when using SCD as the measurement assay for fragmentation percentage. \\\vspace*{0.5cm}

The first step in automating the analysis of sperm cells is to estimate the morphological metrics. According to the WHO Laboratory Manual for the Examination and Processing of Human Semen \cite{world2021laboratory} a sperm head is considered abnormal if the length-to-width ratio is less than 1.5 (round) or larger than 2 (elongated). Additionally, the acrosomal region should comprise 40\% to 70\% of the total sperm head. Recently, Jacobs \cite{jacobs2024image} presented an unsupervised method for objectively measuring the acrosomal region, a task which is typically subjective, and susceptible to high inter-observer variance. \\\vspace*{0.5cm}

A comprehensive review and discussion around clinical guidelines for DNA fragmentation analysis is presented by Agrawal et al. \cite{agarwal2020sperm}. Therein, the authors conduct a SWOT (strength, weakness, opportunity, threat) analysis as well as a comparative table of SDF tests. With this context, there is a clear need for an assistive technology which allows for the selection of a candidate sperm for artificial fertilisation without chemicals or other destructive behaviour. Based on the previous analysis and correlations found between semen parameters and DNA fragmentation levels, this work aims to develop an array of machine learning and image processing based models which automatically assess the semen parameters from bright field and phase contrast images and, using this assessment, predict the level of DNA fragmentation present in the imaged of sperm cell. A broader discussion on recent advances in sperm analysis is presented by Dai et al. \cite{dai2021advances}.\\\vspace*{0.5cm}

Recently, machine learning has become ubiquitous in many facets of modern life and applications in health care have shown tremendous progress. Particularly in the field of andrology, researchers have applied an array of machine learning tools to the problem of sperm analysis. In \cite{mccallum2019deep} the authors present a deep learning approach (pretrained VGG16 convolutional neural network) to predicting a sperm cell's DNA fragmentation level based on the AO chemical assay resulting in a moderate correlation (bivariate correlation of 0.43) between a sperm cell image and DNA quality. Noy et al. \cite{noy2023sperm} propose a model based on quantitative phase imaging of sperm with the model predicting both the fragmentation level as well as the confidence of that prediction. The ground truth fragmentation level of their data is also based on the AO test. Additionally Wang et al. \cite{wang2019prediction} use linear and nonlinear (logistic) regression techniques to predict DNA integrity finding significant correlation between morphological parameters and DNA integrity, again using data collected from the AO test. \\\vspace*{0.5cm}

A hybrid approach to measuring morphometric properties of sperm is presented in \cite{bijar2012fully}, where the authors illustrate how a combination of image processing techniques, some based on partial differential equations, Bayesian classification and a variation in local entropy method for tail identification, are able to accurately segment the components of sperm cells.\\\vspace*{0.5cm}

To mitigate the effects of subjectivity inherent in manual assessment, Kuroda et al. \cite{kuroda2023novel} present a novel AI based framework for the assessment of the ``halos" produced by SCD chemical assays (GoldCyto and HaloSperm).\\\vspace*{0.5cm}

A comprehensive survey of AI applications in andrology is presented by Selvam et al. \cite{panner2024current}. The applications range from predicting sperm DNA integrity to predicting semen parameters from environmental factors (as in \cite{girela2013semen}). Research in applying machine learning to classifying sperm morphology has also been undertaken. As in \cite{spencer2022ensembled}, the authors show how stacking a number of pretrained convolutional neural network architectures into an ensemble model can improve the classification accuracy when trained on the HuSHeM \cite{shaker2017dictionary} and SCIAN \cite{chang2017gold} datasets.\\\vspace*{0.5cm}

In spite of the tremendous advancements already made in AI applications to semen analysis, there are still many challenges to overcome to bridge the gap from science to clinical practice. Riegler et al. \cite{riegler2021artificial} present a candid review of proposed methodologies with a particular focus on clinical relevance. \\\vspace*{0.5cm}

The methodology proposed in this work bears the clinical implementation front of mind and holds the ultimate use case of sperm selection for ICSI or IVF paramount. We aim to achieve this with maximal accuracy of the trained models, and the development of a user friendly software interface which runs machine learning inferences for the host of assays trained on a live video feed of semen samples. The present work illustrates a novel hybrid model which combines morphological measurements known to correlate with sperm DNA fragmentation, as well as state-of-the-art machine learning techniques to enable optimal performance across all chemical assay data.
%% Problem Description
%based on the above, can we train a ml based framework to predict DNA frag from morphological parameters.
%how do we do that? unstained -> DNA, DQ -> DNA  as a validation metric

%% Our contribution
%Ensemble approach
%leverage trinarization paper
%no other study has been so comprehensive
%present correlation between tests? I think for the other paper.

%% file: tex/methods.tex
\section{Methodology}
Computer vision, and specifically image classification, was largely dominated by convolutional neural networks (CNN) and showed tremendous efficacy. AlexNet \cite{krizhevsky2012imagenet} marked an advent for larger and more powerful CNN architectures (VGG \cite{simonyan2014very}, ResNet \cite{he2016deep}, etc.). More recently transformers \cite{vaswani2017attention} have dominated many fields of machine learning, with  computer vision \cite{yin2022vit} being no exception.\\\vspace*{0.5cm}

The seminal work presented by Dosovitskiy et al. \cite{dosovitskiy2020image}, shows how the transformer architecture can be efficiently adapted to image recognition tasks. These technologies are now ubiquitous in the field and many iterations and improvements have been made \cite{han2022survey, khan2022transformers}.\\\vspace*{0.5cm}

A general purpose backbone for computer vision is presented by Liu et al. \cite{liu2021swin} with an update in \cite{liu2022swin}. This new vision transformer presents a hierarchical approach using shifted windows (SWIN) which enables the ability to model at various scales. Hatamizadeh \cite{hatamizadeh2023global} present another iteration of the vision transformer technology which introduces global context (GCVIT) which efficiently models long and short range spatial interactions without the need of shifting windows. For this reason we employ GCVIT as the model backbone for our proposed work.\\\vspace*{0.5cm}

\begin{figure}[h!]
	\centering
	  \includegraphics[width=0.65\linewidth]{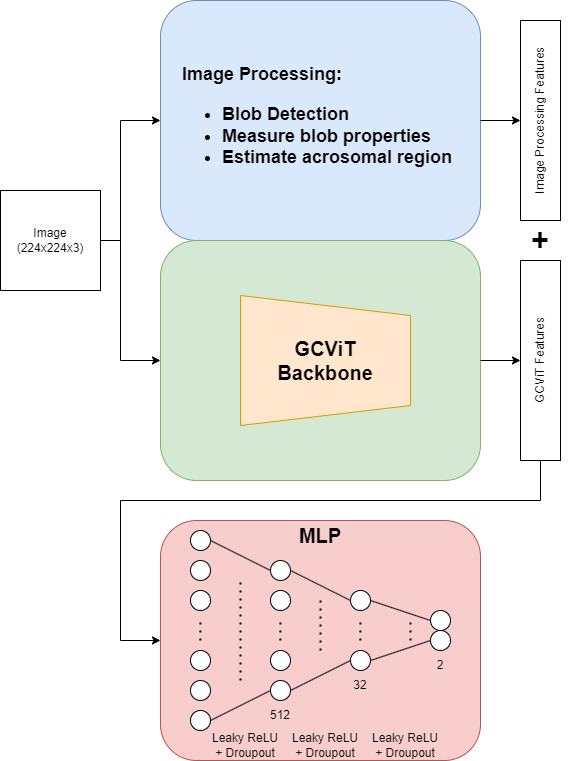}
	  \caption{Schematic of ensemble model architecture for binary classification problem.}
	  \label{fig:architecture}
\end{figure}

In effect we employ a GCVIT architecture, pre-trained on the ImageNet-1k dataset \cite{deng2009imagenet}, as a feature extractor to learn a concise representation of the training data. This pre-trained model is fine tuned on our domain specific dataset, a process termed transfer learning \cite{weiss2016survey}. In conjunction with the features extracted by the transformer model, the image features are measured using image processing techniques. The reason for employing hand selected features is due to the clear WHO guidelines \cite{world2021laboratory} regarding the morphological properties of healthy sperm, as well as the well established correlation between normal morphology and low DNA fragmentation. The learned and selected features are then fed into a multi-layer perceptron (MLP) for final classification or regression. \\\vspace*{0.5cm}
\textbf{Figure \ref{fig:architecture}} presents a schematic layout of the ensemble model proposed in this work. The input image is presented to an image processing pipeline which produces a number of image-based features. These features are:
\begin{itemize}
	\item area ($A$),
	\item major axis length,
	\item minor axis length,
	\item acrosomal region (via \cite{jacobs2024image}),
	\item eccentricity,
	\item equivalent diameter area,
	\item perimeter ($P$),
	\item circularity ($C$),
\end{itemize}
where circularity is calculated as
\begin{equation}
	C = \frac{4 \pi A}{P^2}.
\end{equation}
Additionally the input image is presented to the transformer backbone, in our case the GCVIT, as an ML based feature extractor. The extracted features from both pipelines are processed by a MLP which makes the ultimate classification.\\\vspace*{0.5cm}

%% file: tex/results.tex
\section{Results and Discussion}
In this section we discuss the validation accuracy of the various machine learning models. Each of the fluorometric chemical assays has model trained on brightfield or phase contrast images at the input and a classification based on the corresponding, per sperm, fluorescent image. The intensity of the fluorescent image is used to indicate a positive for DNA fragmentation (stained by the assay) or negative (unstained) result.\\\vspace*{0.5cm} 

For the Aniline Blue and Toluidine Blue tests, the assay interpretation is based on the intensity of the stain. We use the average intensity of the sperm head as an indicator of fragmentation (either postive or negative). The input images for these assays need to be stripped of colour/intensity information to avoid self referential learning. As such the input images are trinarized \cite{jacobs2024image} which produces a greyscale image, from which morphological parameters may be assessed as well as an estimation of the acrosomal region.\\\vspace*{0.5cm}

	\subsection{Aniline Blue}
\begin{figure}[h!]
			\centering
			\begin{subfigure}{0.45\textwidth}
				\includegraphics[width=\linewidth]{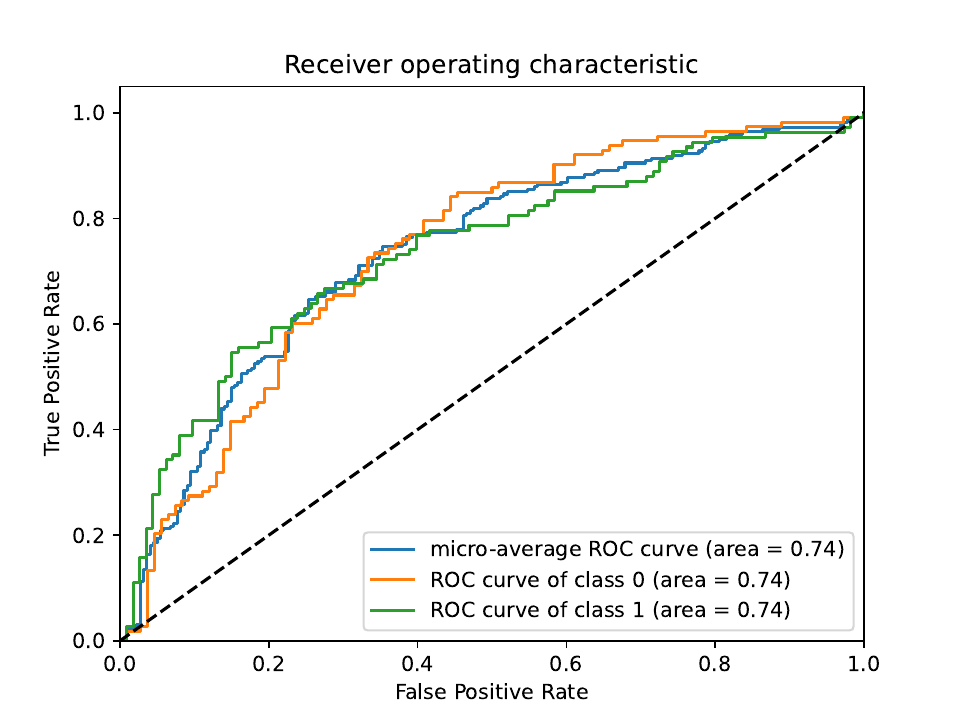}
				\caption{Receiver Operating Characteristic Curve for Aniline Blue with Brightfield Input Images.}
				\label{fig:ROC-AB-BF}
			\end{subfigure}
			\hfill		
			\begin{subfigure}{0.45\textwidth}
				\includegraphics[width=\linewidth]{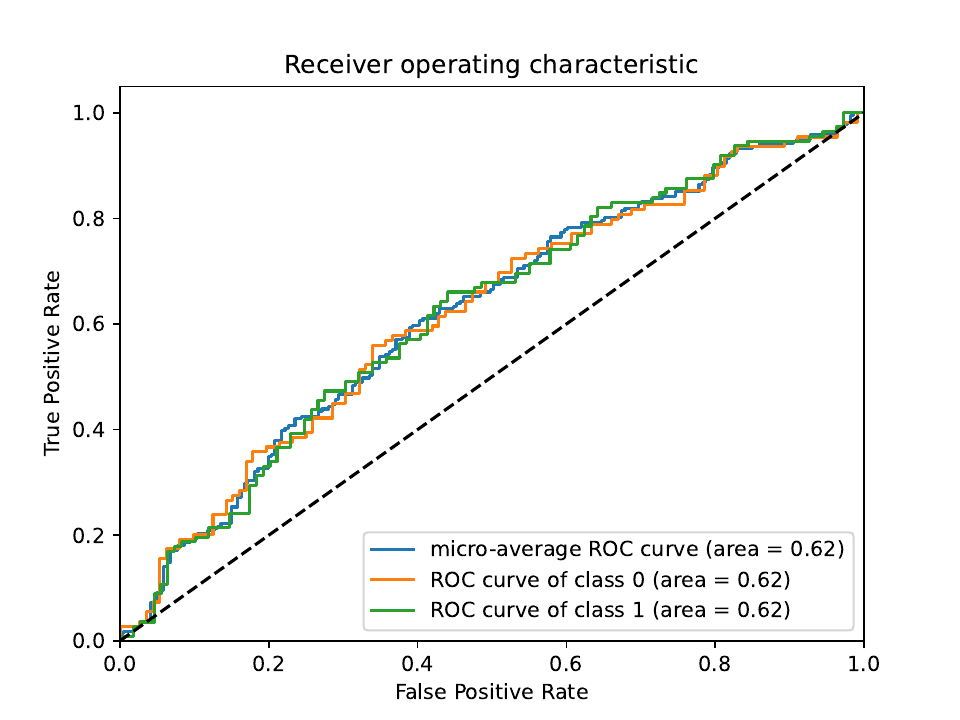}
				\caption{Receiver Operating Characteristic Curve for Aniline Blue with Phase Contrast Input Images.}
				\label{fig:ROC-AB-PC}
			\end{subfigure}
			\hfill
			\caption{Aniline Blue ROC Curves}
			\label{fig:ROC-AB}
		\end{figure}

\begin{figure}[h!]
			\centering
			\begin{subfigure}{0.24\textwidth}
				\includegraphics[width=\linewidth]{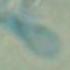}
				\caption{Example image under brightfield.}
				\label{fig:AB-BF-Ex}
			\end{subfigure}
			\hfill
			\begin{subfigure}{0.24\textwidth}
				\includegraphics[width=\linewidth]{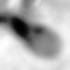}
				\caption{Example image under brightfield after trinarization.}
				\label{fig:AB-BF-Tri-Ex}
			\end{subfigure}
			\hfill		
			\begin{subfigure}{0.24\textwidth}
				\includegraphics[width=\linewidth]{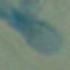}
				\caption{Example image under phase contrast.}
				\label{fig:AB-PC-Ex}
			\end{subfigure}
			\hfill
			\begin{subfigure}{0.24\textwidth}
				\includegraphics[width=\linewidth]{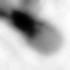}
				\caption{Example image under phase contrast after trinarization.}
				\label{fig:AB-PC-Tri-Ex}
			\end{subfigure}
			\hfill		
			\caption{Aniline Blue Example Images}
			\label{fig:AB-Examples}
		\end{figure}

		The Aniline Blue assay is used to differentiate between mature and immature sperm cells by staining the nuclear lysine-rich histone proteins in immature sperm cells dark blue. Mature sperm with protamines (cysteine- and arginine-rich) appear lighter in colour with a slight blue hue \cite{terquem1983aniline}. The following presents classification reports for our machine learning using both brightfield (Table \ref{tab:AB-BF-Class-Report}) and phase contrast images (Table \ref{tab:AB-PC-Class-Report}) at the input. Furthermore we present the Receiver Operating Characteristic (ROC) curves for the aforementioned test cases in \textbf{Figure \ref{fig:ROC-AB}}. Example images for the Aniline Blue test are shown in \textbf{Figure \ref{fig:AB-Examples}}. The results show fair performance of the model, with the phase contrast model out performing the brightfield model, likely due to the clearer visualisation of the acrosomal region.
		\begin{table}[h!]
			\centering
			\caption{Classification Report for Aniline Blue with Brightfield input.}
			\label{tab:AB-BF-Class-Report}
			\begin{tabular}{|c|c|c|c|c|}
				\hline
				Class & Precision & Recall & F1-Score & Support \\
				\hline
				Unfragmented & 0.66 & 0.77 & 0.71 & 113 \\
				Fragmented & 0.71 & 0.59 & 0.65 & 108 \\
				\hline
				Accuracy & & & 0.68 & 221 \\
				Macro Avg & 0.69 & 0.68 & 0.68 & 221 \\
				Weighted Avg & 0.69 & 0.68 & 0.68 & 221 \\
				\hline
			\end{tabular}
	
		\end{table}
	
		\begin{table}[h!]
			\centering
			\caption{Classification Report for Aniline Blue with Phase Contrast input.}
			\label{tab:AB-PC-Class-Report}
			\begin{tabular}{|c|c|c|c|c|}
				\hline
				Class & Precision & Recall & F1-Score & Support \\
				\hline
				Unfragmented & 0.59 & 0.58 & 0.59 & 109 \\
				Fragmented & 0.60 & 0.62 & 0.61 & 112 \\
				\hline
				Accuracy & & &  0.60 & 221 \\
				Macro Avg & 0.60 & 0.60 & 0.60 & 221 \\
				Weighted Avg & 0.60 & 0.60 & 0.60 & 221 \\
				\hline
			\end{tabular}
		\end{table}
	
	\newpage
	\subsection{Acridine Orange}
\begin{figure}[h!]
			\centering
			\begin{subfigure}{0.45\textwidth}
				\includegraphics[width=\linewidth]{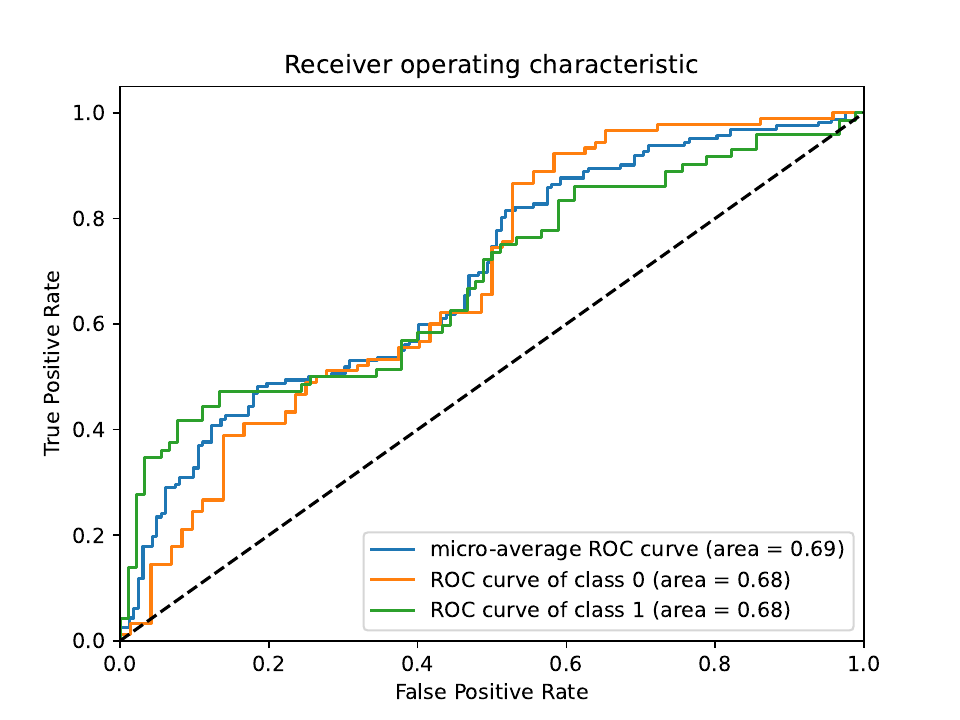}
				\caption{Receiver Operating Characteristic Curve for Acridine Orange with Brightfield Input Images.}
				\label{fig:ROC-AO-BF}
			\end{subfigure}
			\hfill		
			\begin{subfigure}{0.45\textwidth}
				\includegraphics[width=\linewidth]{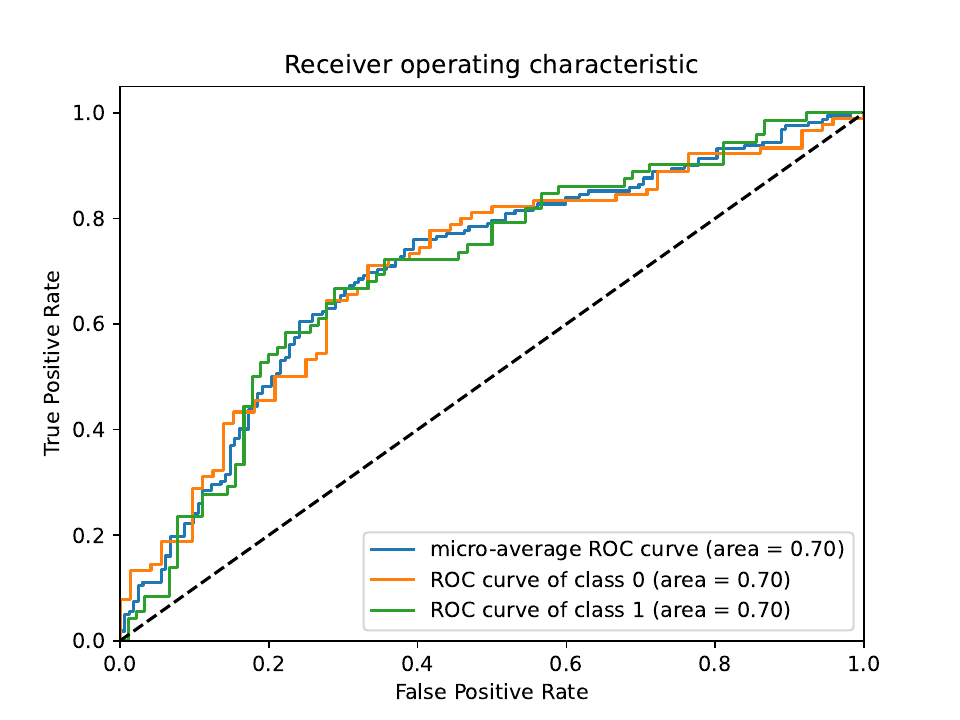}
				\caption{Receiver Operating Characteristic Curve for Acridine Orange with Phase Contrast Input Images.}
				\label{fig:ROC-AO-PC}
			\end{subfigure}
			\hfill
			\caption{Acridine Orange ROC Curves}
			\label{fig:ROC-AO}
		\end{figure}

\begin{figure}[h!]
			\centering
			\begin{subfigure}{0.24\textwidth}
				\includegraphics[width=\linewidth]{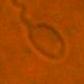}
				\caption{Example image under brightfield.}
				\label{fig:AO-BF-Ex}
			\end{subfigure}
			\hfill
			\begin{subfigure}{0.24\textwidth}
				\includegraphics[width=\linewidth]{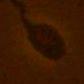}
				\caption{Example image under phase contrast.}
				\label{fig:AO-PC-Ex}
			\end{subfigure}
			\hfill		
			\begin{subfigure}{0.24\textwidth}
				\includegraphics[width=\linewidth]{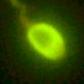}
				\caption{Example image under fluorescence.}
				\label{fig:AO-FL-Ex}
			\end{subfigure}
			\hfill		
			\caption{Acridine Orange Example Images}
			\label{fig:AO-Examples}
		\end{figure}

		Acridine Orange is a fluorescent assay useful in the assessment of sperm DNA integrity, differentiating between normal double-stranded DNA and denatured single-stranded DNA using the properties of the stain \cite{tejada1984test}. The classification reports obtained are presented in Table \ref{tab:AO-BF-Class-Report} and Table \ref{tab:AO-PC-Class-Report} for brightfield and  phase contrast images, respectively. The ROC curves are shown in \textbf{Figure \ref{fig:ROC-AO}} with example images for the Acridine Orange test in \textbf{Figure \ref{fig:AO-Examples}}. Again the model exhibits fair performance performing better on phase contrast images.
		\begin{table}[h!]
			\centering
			\caption{Classification Report for Acridine Orange with Brightfield input.}
			\label{tab:AO-BF-Class-Report}
				\begin{tabular}{|c|c|c|c|c|}
					\hline
					Class & Precision & Recall & F1-Score & Support \\
					\hline
					Unfragmented  & 0.64 & 0.62 & 0.63 & 90 \\
					Fragmented & 0.55 & 0.57 & 0.56 & 72 \\
					\hline
					Accuracy & & &  0.60 & 162  \\
					Macro Avg & 0.60 & 0.60 & 0.60 & 162 \\
					Weighted Avg & 0.60 & 0.60 & 0.60 & 162 \\
					\hline
				\end{tabular}			
		\end{table}
		
		\begin{table}[h!]
			\centering
			\caption{Classification Report for Acridine Orange with Phase Contrast input.}
			\label{tab:AO-PC-Class-Report}
				\begin{tabular}{|c|c|c|c|c|}
					\hline
					Class & Precision & Recall & F1-Score & Support \\
					\hline
					0 & 0.71 & 0.71 & 0.71 & 90 \\
					1 & 0.64 & 0.64 & 0.64 & 72 \\
					\hline
					Accuracy & & & 0.68 & 162 \\
					Macro Avg & 0.68 & 0.68 & 0.68 & 162 \\
					Weighted Avg & 0.68 & 0.68 & 0.68 & 162 \\
					\hline
				\end{tabular}			
		\end{table}
		
	\newpage
	\subsection{CMA3}
\begin{figure}[h!]
			\centering
			\begin{subfigure}{0.45\textwidth}
				\includegraphics[width=\linewidth]{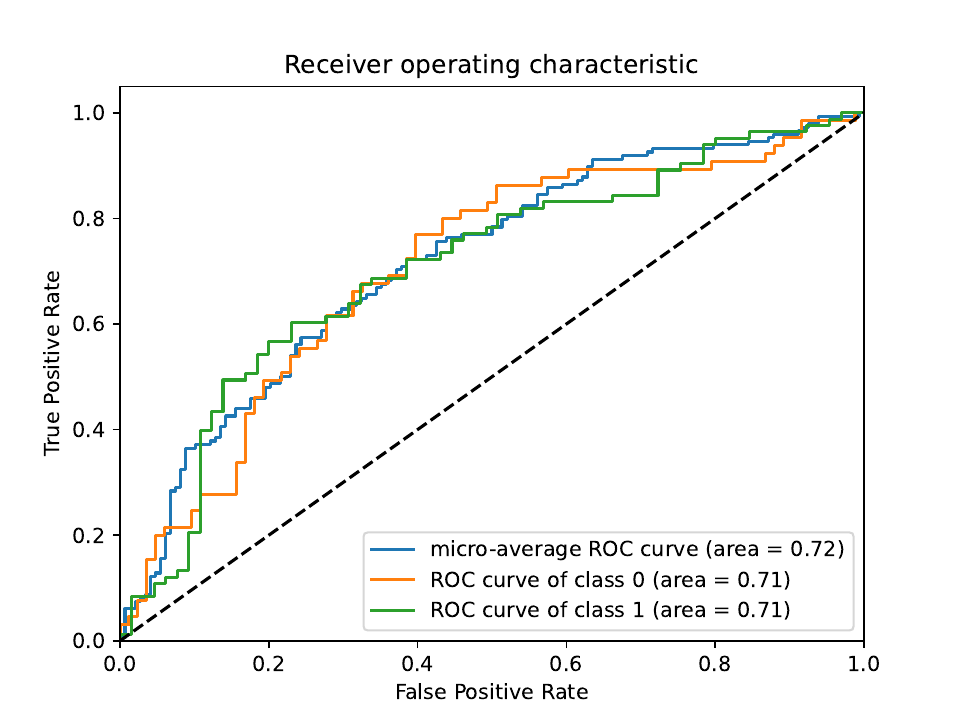}
				\caption{Receiver Operating Characteristic Curve for CMA3 with Brightfield Input Images.}
				\label{fig:ROC-CMA3-BF}
			\end{subfigure}
			\hfill		
			\begin{subfigure}{0.45\textwidth}
				\includegraphics[width=\linewidth]{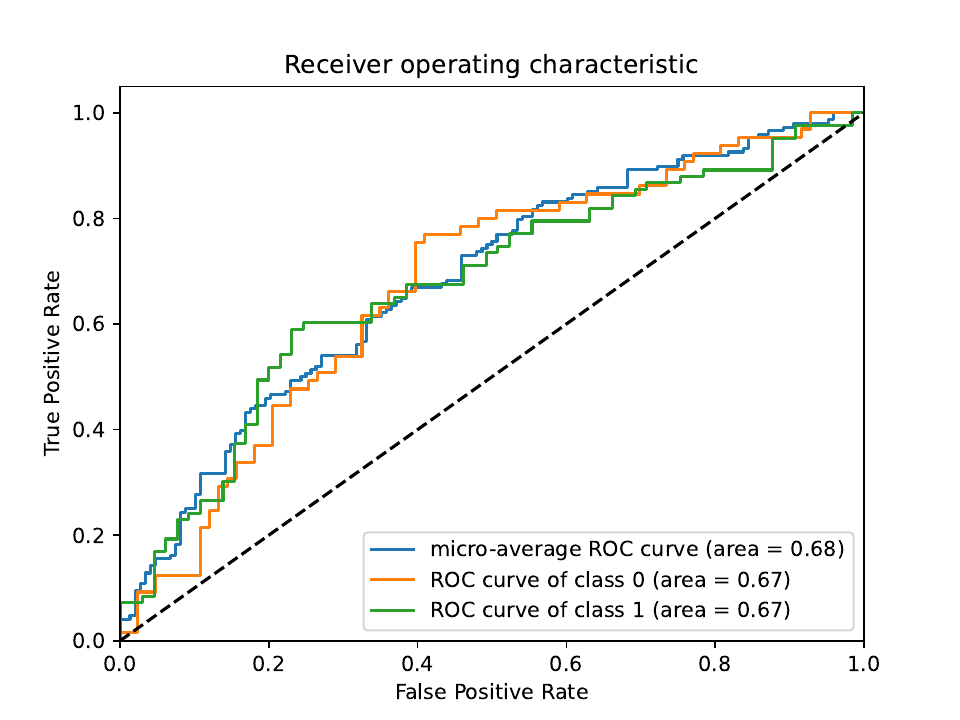}
				\caption{Receiver Operating Characteristic Curve for CMA3 with Phase Contrast Input Images.}
				\label{fig:ROC-CMA3-PC}
			\end{subfigure}
			\hfill
			\caption{CMA3 ROC Curves}
			\label{fig:ROC-CMA3}
		\end{figure}

\begin{figure}[h!]
			\centering
			\begin{subfigure}{0.24\textwidth}
				\includegraphics[width=\linewidth]{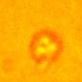}
				\caption{Example image under brightfield.}
				\label{fig:CMA3-BF-Ex}
			\end{subfigure}
			\hfill
			\begin{subfigure}{0.24\textwidth}
				\includegraphics[width=\linewidth]{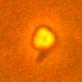}
				\caption{Example image under phase contrast.}
				\label{fig:CMA3-PC-Ex}
			\end{subfigure}
			\hfill		
			\begin{subfigure}{0.24\textwidth}
				\includegraphics[width=\linewidth]{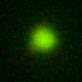}
				\caption{Example image under fluorescence.}
				\label{fig:CMA3-FL-Ex}
			\end{subfigure}
			\hfill		
			\caption{CMA3 Example Images}
			\label{fig:CMA3-Examples}
		\end{figure}

		Similar to the Aniline Blue assay, the CMA3 assay detects protamine deficiency in sperm cells, allowing for the classification of mature/immature cells \cite{lolis1996chromomycin}.  Tables \ref{tab:CMA3-BF-Class-Report} and \ref{tab:CMA3-PC-Class-Report} present the classification reports for brightfield and phase contrast images. \textbf{Figure \ref{fig:ROC-CMA3}} presents ROC curves and \textbf{Figure \ref{fig:CMA3-Examples}} presents example images for CMA3. The model exhibits comparable performance on either image type.
		\begin{table}[h!]
			\centering
			\caption{Classification Report for CMA3 with Brightfield input.}
			\label{tab:CMA3-BF-Class-Report}
			\begin{tabular}{|c|c|c|c|c|}
				\hline
				Class & Precision & Recall & F1-Score & Support \\
				\hline
				Unfragmented & 0.63 & 0.51 & 0.56 & 65 \\
				Fragmented & 0.67 & 0.77 & 0.72 & 83 \\
				\hline
				Accuracy & & & 0.66 & 148 \\
				Macro Avg & 0.65 & 0.64 & 0.64 & 148 \\
				Weighted Avg & 0.65 & 0.66 & 0.65 & 148 \\
				\hline
			\end{tabular}			
		\end{table}
		
		\begin{table}[h!]
			\centering
			\caption{Classification Report for CMA3 with Phase Contrast input.}
			\label{tab:CMA3-PC-Class-Report}
			\begin{tabular}{|c|c|c|c|c|}
				\hline
				Class & Precision & Recall & F1-Score & Support \\
				\hline
				Unfragmented & 0.61 & 0.48 & 0.53 & 65 \\
				Fragmented & 0.65 & 0.76 & 0.70 & 83 \\
				\hline
				Accuracy & & &  0.64 & 148  \\
				Macro Avg & 0.63 & 0.62 & 0.62 & 148 \\
				Weighted Avg & 0.63 & 0.64 & 0.63 & 148 \\
				\hline
			\end{tabular}
		\end{table}
		
	\newpage
	\subsection{Toluidine Blue}
\begin{figure}[h!]
		\centering
		\begin{subfigure}{0.45\textwidth}
			\includegraphics[width=\linewidth]{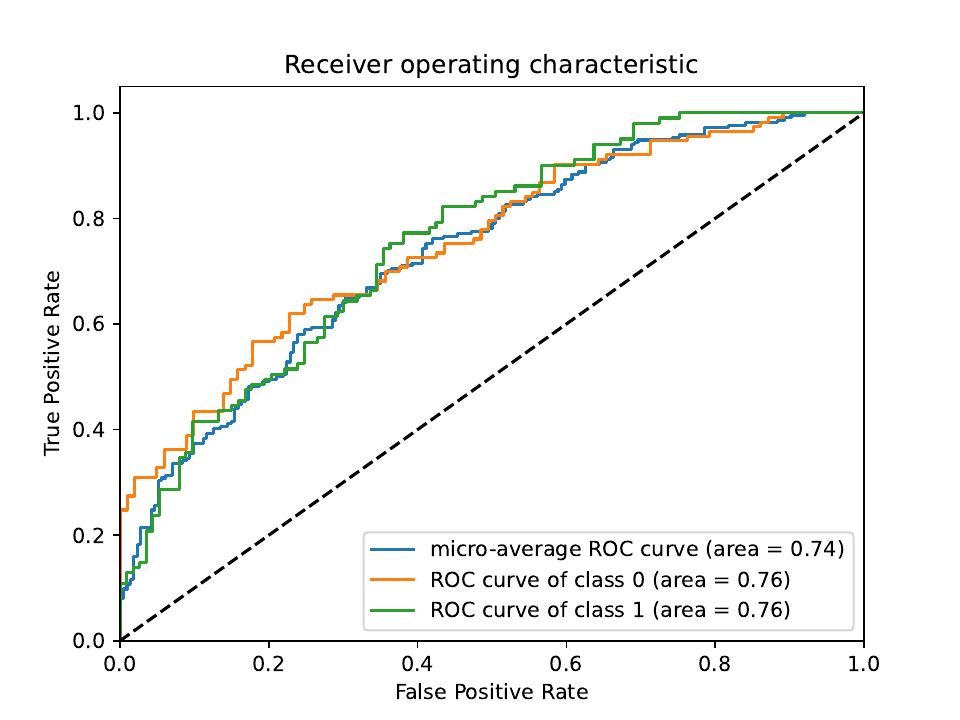}
			\caption{Receiver Operating Characteristic Curve for Toluidine Blue with Brightfield Input Images.}
			\label{fig:ROC-TB-BF}
		\end{subfigure}
		\hfill		
		\begin{subfigure}{0.45\textwidth}
			\includegraphics[width=\linewidth]{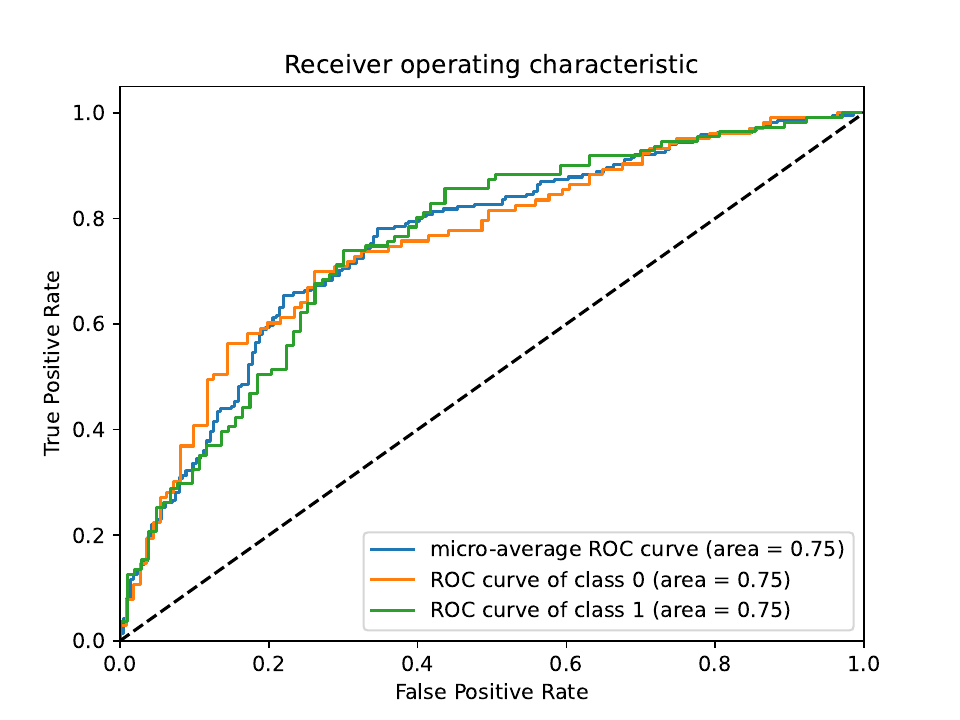}
			\caption{Receiver Operating Characteristic Curve for Toluidine Blue with Phase Contrast Input Images.}
			\label{fig:ROC-TB-PC}
		\end{subfigure}
		\hfill
		\caption{Toluidine Blue ROC Curves}
		\label{fig:ROC-TB}
	\end{figure}

\begin{figure}[h!]
		\centering
		\begin{subfigure}{0.24\textwidth}
			\includegraphics[width=\linewidth]{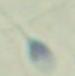}
			\caption{Example image under brightfield.}
			\label{fig:TB-BF-Ex}
		\end{subfigure}
		\hfill
		\begin{subfigure}{0.24\textwidth}
			\includegraphics[width=\linewidth]{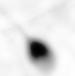}
			\caption{Example image under brightfield after trinarization.}
			\label{fig:TB-BF-Tri-Ex}
		\end{subfigure}
		\hfill		
		\begin{subfigure}{0.24\textwidth}
			\includegraphics[width=\linewidth]{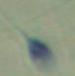}
			\caption{Example image under phase contrast.}
			\label{fig:TB-PC-Ex}
		\end{subfigure}
		\hfill
		\begin{subfigure}{0.24\textwidth}
			\includegraphics[width=\linewidth]{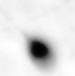}
			\caption{Example image under phase contrast after trinarization.}
			\label{fig:TB-PC-Tri-Ex}
		\end{subfigure}
		\hfill		
		\caption{Toluidine Blue Example Images}
		\label{fig:TB-Examples}
	\end{figure}

	Toluidine Blue is a sensitive test for incomplete DNA structure and packaging \cite{talebi2012cytochemical}. Sperm cells with good chromatin structure stain light blue while those with abnormal chromatin structure stain dark blue. Tables \ref{tab:TB-BF-Class-Report} and \ref{tab:TB-PC-Class-Report} present the classification reports for brightfield and phase contrast images. \textbf{Figure \ref{fig:ROC-TB}} presents ROC curves and \textbf{Figure \ref{fig:TB-Examples}} presents example images for the Toluidine Blue test. The model exhibits slightly better performance on phase contrast images.
	\begin{table}[h!]
		\centering
		\caption{Classification Report for Toluidine Blue with Brightfield input.}
		\label{tab:TB-BF-Class-Report}
		\begin{tabular}{|c|c|c|c|c|}
			\hline
			Class & Precision & Recall & F1-Score & Support \\
			\hline
			Unfragmented & 0.78 & 0.51 & 0.62 & 113 \\
			Fragmented & 0.61 & 0.84 & 0.71 & 101 \\
			\hline
			Accuracy & & & 0.67 & 214 \\
			Macro Avg & 0.70 & 0.68 & 0.66 & 214 \\
			Weighted Avg & 0.70 & 0.67 & 0.66 & 214 \\
			\hline
		\end{tabular}

	\end{table}
	
	\begin{table}[h!]
		\centering
		\caption{Classification Report for Toluidine Blue with Phase Contrast input.}
		\label{tab:TB-PC-Class-Report}
		\begin{tabular}{|c|c|c|c|c|}
			\hline
			Class & Precision & Recall & F1-Score & Support \\
			\hline
			Unfragmented & 0.73 & 0.60 & 0.66 & 103 \\
			Fragmented & 0.68 & 0.79 & 0.73 & 111 \\
			\hline
			Accuracy & & & 0.70 & 214 \\
			Macro Avg & 0.71 & 0.70 & 0.70 & 214 \\
			Weighted Avg & 0.70 & 0.70 & 0.70 & 214 \\
			\hline
		\end{tabular}
	\end{table}
	
	\newpage
	\subsection{TUNEL}
\begin{figure}[h!]
			\centering
			\begin{subfigure}{0.45\textwidth}
				\includegraphics[width=\linewidth]{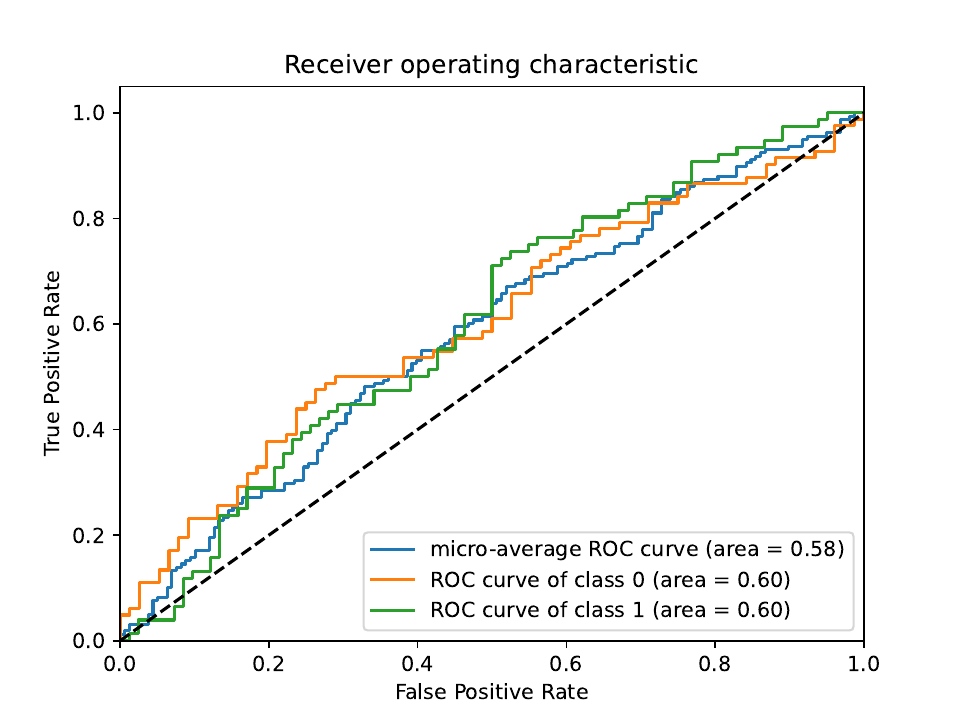}
				\caption{Receiver Operating Characteristic Curve for TUNEL with Brightfield Input Images.}
				\label{fig:ROC-TUNEL-BF}
			\end{subfigure}
			\hfill		
			\begin{subfigure}{0.45\textwidth}
				\includegraphics[width=\linewidth]{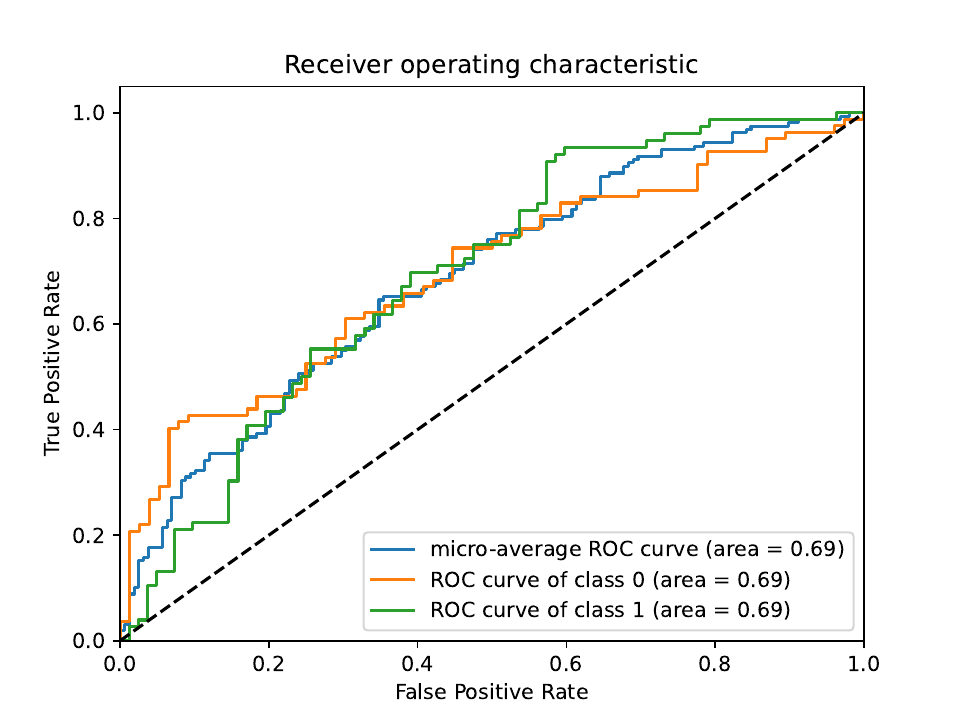}
				\caption{Receiver Operating Characteristic Curve for TUNEL with Phase Contrast Input Images.}
				\label{fig:ROC-TUNEL-PC}
			\end{subfigure}
			\hfill
			\caption{TUNEL ROC Curves}
			\label{fig:ROC-TUNEL}
		\end{figure}

\begin{figure}[h!]
			\centering
			\begin{subfigure}{0.2\textwidth}
				\includegraphics[width=\linewidth]{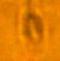}
				\caption{Example image under brightfield.}
				\label{fig:TUNEL-BF-Ex}
			\end{subfigure}
			\hfill
			\begin{subfigure}{0.2\textwidth}
				\includegraphics[width=\linewidth]{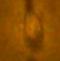}
				\caption{Example image under phase contrast.}
				\label{fig:TUNEL-PC-Ex}
			\end{subfigure}
			\hfill		
			\begin{subfigure}{0.2\textwidth}
				\includegraphics[width=\linewidth]{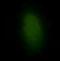}
				\caption{Example image under fluorescence.}
				\label{fig:TUNEL-FL-Ex}
			\end{subfigure}
			\hfill		
			\caption{TUNEL Example Images}
			\label{fig:TUNEL-Examples}
		\end{figure}

		Finally, the widely used clinical test, TUNEL, is considered. Fluorescent images with a bright luminosity indicate sperm cells with a high degree of DNA fragmentation. Table \ref{tab:TUNEL-BF-Class-Report} shows excellent classification performance on the validation set, while the phase contrast counterpart also performs well (Table \ref{tab:TUNEL-PC-Class-Report}). We attribute this performance to the clear indication of DNA fragmentation in the flourescent images. Example images are shown in \textbf{Figure \ref{fig:TUNEL-Examples}}. Additionally, ROC curves shown in \textbf{Figure \ref{fig:ROC-TUNEL}} report area under curve (AUC) values of 0.82 and 0.80 for brightfield and phase contrast respectively. The micro-average curve aggregates the contributions from both and computes the average metrics.
%		\begin{table}[h!]
%			\centering
%			\caption{Classification Report for TUNEL with Brightfield input.}
%			\label{tab:TUNEL-BF-Class-Report}
%
%			\begin{tabular}{|c|c|c|c|c|}
%				\hline
%				Class & Precision & Recall & F1-Score & Support \\
%				\hline
%				Unfragmented & 0.80 & 0.80 & 0.80 & 5 \\
%				Fragmented & 0.93 & 0.93 & 0.93 & 14 \\
%				\hline
%				Accuracy & & & 0.89 & 19 \\
%				Macro Avg & 0.86 & 0.86 & 0.86 & 19 \\
%				Weighted Avg & 0.89 & 0.89 & 0.89 & 19 \\
%				\hline
%			\end{tabular}
%		
%		\end{table}
%		
%		\begin{table}[h!]
%			\centering
%			\caption{Classification Report for TUNEL with Phase Contrast input.}
%			\label{tab:TUNEL-PC-Class-Report}
%
%			\begin{tabular}{|c|c|c|c|c|}
%				\hline
%				Class & Precision & Recall & F1-Score & Support \\
%				\hline
%				Unfragmented  & 0.75 & 0.60 & 0.67 & 5 \\
%				Fragmented  & 0.87 & 0.93 & 0.90 & 14 \\
%				\hline
%				Accuracy & & & 0.84 & 19 \\
%				Macro Avg & 0.81 & 0.76 & 0.78 & 19 \\
%				Weighted Avg & 0.84 & 0.84 & 0.84 & 19 \\
%				\hline
%			\end{tabular}
%		
%		\end{table}
		\begin{table}[h!]
			\centering
			\caption{Classification Report for TUNEL with Brightfield input.}
			\label{tab:TUNEL-BF-Class-Report}
			
			\begin{tabular}{|c|c|c|c|c|}
				\hline
				Class & Precision & Recall & F1-Score & Support \\
				\hline
				Unfragmented & 0.65 & 0.34 & 0.45 & 82 \\
				Fragmented & 0.53 & 0.80 & 0.64 & 76 \\
				\hline
				Accuracy & & & 0.56 & 158 \\
				Macro Avg & 0.59 & 0.57 & 0.54 & 158 \\
				Weighted Avg & 0.59 & 0.56 & 0.54 & 158 \\
				\hline
			\end{tabular}
			
		\end{table}
		
		\begin{table}[h!]
			\centering
			\caption{Classification Report for TUNEL with Phase Contrast input.}
			\label{tab:TUNEL-PC-Class-Report}
			
			\begin{tabular}{|c|c|c|c|c|}
				\hline
				Class & Precision & Recall & F1-Score & Support \\
				\hline
				Unfragmented  & 0.64 & 0.73 & 0.68 & 82 \\
				Fragmented  & 0.66 & 0.55 & 0.60 & 76 \\
				\hline
				Accuracy & & & 0.65 & 158 \\
				Macro Avg & 0.65 & 0.64 & 0.64 & 158 \\
				Weighted Avg & 0.65 & 0.65 & 0.64 & 158 \\
				\hline
			\end{tabular}
			
		\end{table}
		
		\newpage

%% file: tex/conclusion.tex
\section{Conclusion}
This work presents a comprehensive ensemble framework which combines mathematical analysis, image processing and machine learning to accurately predict the classification of DNA fragmentation of a sperm cell from image data alone. \\\vspace*{0.5cm}
A varied collection of chemical assays was administered to 65 patient samples with image data of the resulting sperm collected, under both bright field and phase contrast (and fluorescence where applicable). \\\vspace*{0.5cm}
The results presented herein indicate fair performance on some assays, due to the subjective indication of DNA fragmentation. However, the results obtained for the TUNEL assay illustrate an excellent capability in predicting the class of DNA fragmentation.\\\vspace*{0.5cm}
This non-destructive approach has the potential to significantly impact the field of reproductive medicine by allowing the accurate prediction of a sperm cell's DNA fragmentation status at the point of ICSI. Future work aims to quantify the present framework's impact of embryo quality and development, and ultimately birth rates of assisted reproduction cycles. The integration of machine learning and image processing in this context opens new avenues for enhancing ART outcomes and addressing the growing challenges of male infertility in a globally declining fertility landscape.